\documentclass[journal]{IEEEtran} 

\usepackage{subcaption}
\usepackage[utf8]{inputenc}
\usepackage[T1]{fontenc}
\usepackage{multirow}\usepackage{booktabs}
\usepackage{siunitx}
\usepackage[colorlinks=true, allcolors=blue]{hyperref}
\usepackage{amsmath}
\usepackage{amsfonts}
\usepackage{soul} 
\usepackage{stfloats}
\usepackage{silence}
\usepackage[font=small,skip=10pt,compatibility=false]{caption}

\usepackage{tikz,xcolor}
\definecolor{lime}{HTML}{A6CE39}
\DeclareRobustCommand{\orcidicon}
{
    \begin{tikzpicture}
    \draw[lime, fill=lime] (0,0) circle [radius=0.16] 
    node[white] {{\fontfamily{qag}\selectfont \tiny ID}};    \draw[white, fill=white] (-0.0625,0.095) circle [radius=0.007];    
    \end{tikzpicture}
    \hspace{0mm}}
\foreach \x in {A, ..., Z}{%
    \expandafter\xdef\csname orcid\x\endcsname{\noexpand\href{https://orcid.org/\csname orcidauthor\x\endcsname}{\noexpand\orcidicon}}
}

\begin{document}

\title{Active Data Acquisition in Autonomous Driving Simulation}

\author{Jianyu Lai, Zexuan Jia, Boao Li
\thanks{Jianyu Lai, Zexuan Jia, Boao Li are with the Department of Computer Science and Technology, Southern University of Science and Technology, Shenzhen, 518055, China}
}

\markboth{Journal of \LaTeX\ Class Files,~Vol.
}%
{Shell \MakeLowercase{\textit{et al.}}: Bare Demo of IEEEtran.cls for IEEE Journals}

\maketitle

\begin{abstract}
Autonomous driving algorithms rely heavily on learning-based models, which require large datasets for training. 
However, there is often a large amount of redundant information in these datasets, while collecting and processing these datasets can be time-consuming and expensive. 
To address this issue, this paper proposes the concept of an active data-collecting strategy. 
For high-quality data, increasing the collection density can improve the overall quality of the dataset, ultimately achieving similar or even better results than the original dataset with lower labeling costs and smaller dataset sizes. 
In this paper, we design experiments to verify the quality of the collected dataset and to demonstrate this strategy can significantly reduce labeling costs and dataset size while improving the overall quality of the dataset, leading to better performance of autonomous driving systems. 
The source code implementing the proposed approach is publicly available on the page \footnote{\url{https://github.com/Th1nkMore/carla_dataset_tools}}.
\end{abstract}


%
\IEEEpeerreviewmaketitle

\section{Introduction}

Currently, most of the algorithms related to autonomous driving use learning-based models, which typically require a large amount of data as support. 
Therefore, the size of autonomous driving datasets is often extremely large (TB level). 
Collecting these data and using them to train autonomous driving algorithms incurs high costs in terms of both collection time and training time. 
However, in practice, there is often a large amount of redundant information in the large-scale datasets used for autonomous driving. 
This redundant information can only have a small positive impact on the training of algorithms, but the cost of collecting and using this redundant information for training is very high. 
Therefore, our goal is to develop a new data collection tool that can reduce the redundant information in the collected dataset and improve the quality of the dataset, thus increasing the training efficiency of the algorithm and reducing the time and cost of data collection and training by providing the data to the algorithm before training. 

In order to improve the quality of the dataset, the precision of data points and the quantity of data are considered as two metrics. 
It is vital to find a balance between them: for high-quality data, increasing the collection density can improve the overall quality of the dataset, ultimately achieving similar or even better results than the original dataset with lower labelling costs and smaller dataset size.
In this project, all experiments were completed and tested in a simulated scene. First, it is cost-effective. 
Conducting tests in real-world scenarios can be time-consuming and expensive \cite{lan2022time,lan2021learning,lan2021learningneuro}. 
Using simulation scenes can significantly reduce the cost of testing. 
By creating virtual scenarios, researchers can quickly test various driving scenarios and abnormal situations, which can improve testing efficiency. 
Second, it is convenient to do data recording and analysis. 
In simulation scenes, various data such as sensor data and driving trajectories can be easily recorded and analyzed. 
These data can be used to improve the algorithms and performance of automatic-driving systems.

Vehicle detection is an important research topic in the field of autonomous driving, and one of the hotspots in the field of computer vision \cite{lightweight}. 
Because of the strong dependency of the object detection algorithm, the quality of a dataset must be evaluated by a specific algorithm, thus one kind of algorithm must me selected as the criterion for the evaluation of the dataset \cite{lan2019evolving,lan2018real}. 
For autonomous vehicles, the comprehension of the surroundings is of primary importance, and more data and images lead to better performance. 
However, considering the project scale and the time usage, it is necessary to choose a proper algorithm that is not only time-saved but also can be trained well with images at mid-scale. 
Considering the difficulty and the engineering requirement, the detection algorithm YOLO v5 is chosen. 
This series of algorithms will be introduced in the following section.

\section{Related Work}
\label{sec:related_work}

This section only briefly discusses our literal review of several popular object detection algorithms. 
Since object detection requires object localization and object type classification \cite{gao2021neat}, the learned features can be categorized into one-stage and two-stage object detection algorithms \cite{objectdetection}.

One-stage object detection algorithms generate object location and object classification results directly in one stage, so they do not require a region proposal process, which is usually simpler and faster than two-stage detection algorithms \cite{lan2022class}. 
There are two kinds of one-stage detection algorithms, single-shot multi-box detector (SSD) \cite{SSD}
and you only look once (YOLO) \cite{YOLO}. 
The SSD algorithm uses a VGG16 network as the backbone, and SSD-512 as the input version. 
While the Yolo detection algorithm is a real-time object detection algorithm, and the Yolo-v3 is the most popular version because of the model's compact size and ease of application \cite{YOLOv3}, which uses the DarkNet-53 as the backbone. 

The two-stage object detection algorithms need to conduct a region proposal process first and then classify the object in the proposed region \cite{lan2019evolving,lan2018real}. 
The basic algorithm is the R-CNN algorithm.
However, the two-stage construction means more time will be used in the training processing, so the R-CNN algorithm is not a light-weighted algorithm so two advanced algorithms, mask R-CNN and faster R-CNN will be introduced.
The Fast R-CNN uses DCNN(deep CNN) as the backbone, and it employs a region proposal algorithm that reduces the number of object region proposals and improves the region proposal quality to improve the learning speed \cite{FastRCNN}.
The Mask R-CNN enhances the overall detection accuracy and small target detection, which the pool layers affect a lot  \cite{MaskRCNN}.

\begin{table}[ht!] \centering
\begin{tabular}{llllll}  \toprule
Measure & Faster R-CNN & YOLOv3   \\ \hline
TP (True Positives)     & 578       & 751        \\
FP (False Positives)    & 2         & 2          \\
FN (False Negatives)    & 150        & 7          \\
Precision (TPR)         & 99.66\%    & 99.73\%   \\
Sensitivity (recall)    & 79.40\%    & 99.07\%   \\
F1 Score  & 88.38\% & 99.94\% \\
Quality  & 79.17\% &98.81\% \\
Processing Time (Av. in ms)  & 1.39s & 0.057ms \\ \bottomrule
\end{tabular}
\caption{Faster R-CNN and YOLO v3}
\label{YOLOv3}
\end{table}

From Table \ref{YOLOv3} \cite{cardetection}, the Yolo-v3 performs better than the fastest two-stage detection algorithm faster R-CNN in all attributes, which reduces the FN a lot and increase the quality and processing time, which means the FPS(frame per second) is much higher and the mAP is more accurate. And From Table \ref{comparison}, all the Yolo series algorithms and faster R-CNN are compared. 
We can find that the mAP of Yolo-v5 and Yolo-v3 is similar, but Yolo-v5 has about 6 times FPS of Yolo-v3 and 2 times FPS as Yolo-v4, which shows that it is very efficient. Because in our project, the scale of the dataset and the training time both count, so after consideration, the light-weighted and accurate algorithm Yolo-v5 is chosen.

\begin{table}[ht!]
\centering
\scalebox{0.85}{
\begin{tabular}{llllll} 
\toprule
Model               & Size (pixels)         & Test dataset  & mAP (0.5)  & FPS & GPU   \\
\hline
Fast R-CNN          & $600 \times 1000$       & VOC2007       & 70        & 0.5 & Titan X       \\
Faster R-CNN        & $600 \times 1000$      & VOC2007       & 76.4      & 5   & Titan X     \\
YOLOv1              & $446 \times 448$        & VOC2007       & 63.4      & 45  & Titan X   \\
SSD                 & $512 \times 512$        & VOC2007       & 76.8      & 19  & Titan X    \\
YOLOv2              & $544 \times 544$        & MS COCO       & 44.0      & 40  & Titan X    \\
YOLOv3              & $608 \times 608$        & MS COCO       & 57.9      & 20  & Titan X    \\
YOLOv4              & $608 \times 608$       & MS COCO       & 65.7      & 62  & Tesla V100    \\
YOLOv5s             & $640 \times 640$        & MS COCO       & 55.4      & 113  & Tesla V100   \\
\bottomrule
\end{tabular}
}

\caption{The specific comparison among YOLO series algorithms.}
\label{comparison}
\end{table}

For the simulation environment, the CARLA \cite{CARLA} simulator is chosen. 
CARLA is an open-source simulator for autonomous driving research. 
CARLA has been developed from the ground up to support the development, training, and validation of autonomous urban driving systems. 
In addition to open-source code and protocols, CARLA provides open digital assets (urban layouts, buildings, vehicles) that were created for this purpose and can be used freely. 
The simulation platform supports flexible specifications of sensor suites and environmental conditions \cite{lan2016development,lan2016developmentuav,xiang2016uav}. 
CARLA is generally used to study the performance of three approaches to autonomous driving: a classic modular pipeline, an end-to-end model trained via imitation learning, and an end-to-end model trained via reinforcement learning. 
The approaches are evaluated in controlled scenarios of increasing difficulty, and their performance is examined via metrics provided by CARLA, illustrating the platform’s utility for autonomous driving research. 
In addition, the motion flow method can be easily used in the data collected in the CARLA simulation environment, which can assist in the analysis of the trajectory of the cars. 
In CARLA, the target ODD can be specified more easily than in real environments. 
In this paper, all the experiments are conducted in and tested in a CARLA simulation environment.

\section{Methodology}
\label{sec:methodology}

\subsection{Indicator algorithms}
\label{sec:indicator}
In object detection tasks, the quality and quantity of training data are critical factors that affect the performance of the algorithm. 
For autonomous driving cars, understanding the surrounding environment is the most important thing \cite{xu2019online}, the YOLO algorithm which is chosen, the more instance is provided, the better the performance is. 
While increasing the number of instances can lead to better performance, it is not always feasible to collect or annotate a large amount of data. 
In this work, we propose a novel approach that utilizes the information provided by YOLO to generate synthetic data that can be used to improve the performance of the algorithm. 
By analyzing the confidence scores and bounding boxes output by YOLO, we can identify the most informative instances and use them to generate high-quality synthetic data. 
In this section, we present our methodology for data generation and evaluation, and demonstrate the effectiveness of our approach on a benchmark dataset.

\subsection{Data Complexity Design}
What is more, to test the algorithm efficiently, the data collection method and strategies are critical. 
For a specific algorithm, the diversity and the quality will affect the training result. 
For diversity, we use the UQI (Universal Image Quality Index) \cite{UQI} as one of the metrics. 
When the similarity of two consecutive images is higher than one threshold, the image will be removed. 
This method can grant the diversity of the training dataset and prove that no similar images will be used so that the training time will be reduced as well. 
The formula is as follows:
$$Q = \frac{4\sigma_{xy}\bar{x}\bar{y}}{(\sigma^2_x + \sigma^2_y)[(\bar{x}^2 + \bar{y}^2)]} $$

When it comes to quality, there are several factors, but the main two factors that we consider are the number of objects in each image and the occlusion degree of each object in one image. For the first one, when one image has more target objects, the YOLO algorithm can be trained better with the same data scale, because more features can be shown and learned by the algorithm. Then, for the occlusion degree, because the semantic segmentation image classifies all the same kind of objects in the same color, so when two or more objects are near each other, they might be labelled as one big car. 
So reducing or recognizing this kind of image are very critical.

\subsection{Radar Algorithms}
Radar cross-section (RCS) measurements \cite{groundbased}require a well-calibrated radar system. Often, the calibration process is performed by evaluating the received echo signal from a target with a well-known RCS. In a ground-based setup, radar measurements may be significantly affected by the environment in general and by the multipath in particular. 
Also, Driver assistance and safety systems are in high demand by customers and legislators. Those systems are working satisfactorily on highways, but for urban scenarios, a more precise position, orientation and dimension estimation is essential for time-crucial systems. It is assumed that the target vehicle does not drift, and hence the orientation can be used to determine the direction of movement \cite{highresolution}.

\subsection{Experiment Methods}
For the experiment and testing aspect, the main input of this project is the semantic segmentation of images captured through cameras in the Carla simulation environment, along with instance segmentation to assist in judgment and detection. 
Semantic segmentation can automatically classify different types of objects in the scene, facilitating annotation tools to annotate them. 
Instance segmentation is easy to observe and understand, making it easy to check the results. 
In the data of the Carla simulation tool, all cars and traffic lights (objects to be detected) have their own ground truth, i.e., their actual positions in the image. 
The method used in this article is to first use existing labelling tools to annotate the semantic segmentation images, and then use the annotated data to train the YOLO algorithm.
Finally, give the trained YOLO algorithm some raw segmentation data, let it label the detected objects, and then check the results of properties such as IoU \cite{IOU} and mAP50.

Regarding image quality, there are several influencing factors, but the two main factors are the number of objects in each image and the degree of occlusion of each object in each image.
For the first one, when an image has more target objects, using the same data size can better train the YOLO algorithm because the algorithm can display and learn more features. 
As for occlusion, since the semantic segmentation image classifies all objects of the same type as the same colour when two or more objects are close to each other, they may be labeled as larger objects. 
Reducing or identifying such images is critical.

\section{Experimental setup}

In order to evaluate the effectiveness of the active data collection strategy proposed in this work, which may bring a higher unit data value, two experiments were designed and conducted. The first experiment aimed to compare the performance of a model trained on data collected using the active strategy with the performance of a model trained on data collected using a passive strategy while controlling for other variables. The second experiment aimed to compare the performance of a model trained on a fixed amount of data collected using the active strategy with the performance of a model trained on the same amount of data collected using a passive strategy, while also controlling for other variables.

Here is a brief view of the dataset configuration, which also shows the experiment design:
\begin{itemize}
    \item D1: Collect all data passively 
    \item D1-S: Active collect for the same time as D1
    \item D2:Active collect until the same size as D1
    \item V: Dataset used for validate and test
\end{itemize}

\begin{table}[ht!]
\centering
\begin{tabular}{llllll} 
\toprule
Dataset name & Total Frames & ~Map     \\ 
\hline
D1           & 900          & Town02       \\
D1-S         & 579          & Town02           \\
D2           & 900          & Town02         \\
V            & 375          & Town03       \\
\bottomrule
\end{tabular}
\caption{Dataset Configuration}
\end{table}

\subsection{Equivalent Time-samples Experiment}

This experiment was conducted by collecting two datasets, denoted as dataset $\textbf{D1}$ and dataset $\textbf{D1-S}$ respectively. Dataset $\textbf{D1-S}$ was collected using the active data collection strategy proposed in this work, while dataset $\textbf{D1}$ collects all data. The datasets were collected at the same time, with the same map and vehicle number, in order to make the experiment comparative.

The performance of a model trained on dataset $\textbf{D1}$ was then compared with the performance of a model trained on dataset $\textbf{D1-S}$. Specifically, the metrics of the models trained on the two datasets were compared, and the time required to train the models was also recorded. We expect that the metrics of the model trained on dataset $\textbf{D1-S}$ will be similar to the metrics of the model trained on dataset $\textbf{D1}$, but the time required to train the model using dataset $\textbf{D1-S}$ will be significantly less than the time required to train the model using dataset $\textbf{D1}$.

\subsection{Equivalent Size-samples Experiment}
This experiment was conducted by collecting two datasets, denoted as dataset $\textbf{D1}$ and dataset $\textbf{D2}$ respectively. Data set \textbf{D2} uses the active strategy and data set \textbf{D1} not uses the strategy respectively, and changes map and vehicle number to make the experiment comparative, then observes the time and effect of the training model. The datasets were collected at the same size, with the same map and vehicle number, in order to make the experiment comparative.

The performance of a model trained on dataset $\textbf{D1}$ was then compared with the performance of a model trained on dataset $\textbf{D2}$. Specifically, the metrics of the models trained on the two datasets were compared, and the time required to train the models was also recorded. We expect that the metrics of the model trained on dataset $\textbf{D2}$ will be significantly better than the metrics of the model trained on dataset $\textbf{D1}$, and that the time required to train the model using dataset $\textbf{D1}$ will be similar to the time required to train the model using dataset $\textbf{D2}$.

\section{Results}
\label{sec:results}

\begin{table}[ht!] \centering
\begin{tabular}{llllll} \toprule
Dataset name & Vehicle & Training time & mAP &mAP50~95 \\ \hline
D1&  996    & 14min40s    &  	0.508  & 0.391  \\
D1-S& 996 & 11min32s    & 0.492  & 0.365     \\
D2 &  1835  & 16min07s   & 0.647    & 0.525  \\ \bottomrule
\end{tabular}
\caption{Raw Result}
\end{table}

\begin{figure*}[ht!]  \centering
    \includegraphics[width=0.9\textwidth]{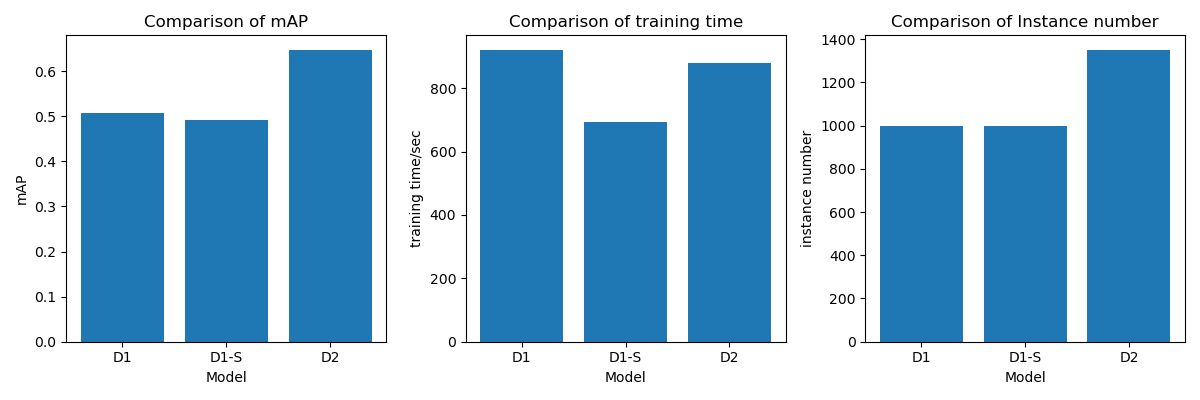}
    \caption{Comparison of mAP, training time and instance number}
    \label{fig:raw result}
\end{figure*}

From the equivalent Time-samples Experiment, we observed that by employing an active acquisition strategy, the number of collected images decreased while the number of instances remained unchanged. The effectiveness of the approach remained similar. However, the training time decreased from 14 minutes and 40 seconds to 11 minutes and 32 seconds (about \textbf{21.36\%} reduction).

From the equivalent Size-samples Experiment, under the same amount of data, the model trained by the dataset obtained using an active acquisition strategy demonstrated a significant increase in the number of instances collected. The training time remained comparable, but there was a noticeable improvement in performance (approximately \textbf{27.36\%} increment).

\begin{figure}[htbp]
    \centering
    \begin{subfigure}[b]{0.45\textwidth}
        \includegraphics[width=\linewidth,trim={0 0 100 50},clip]{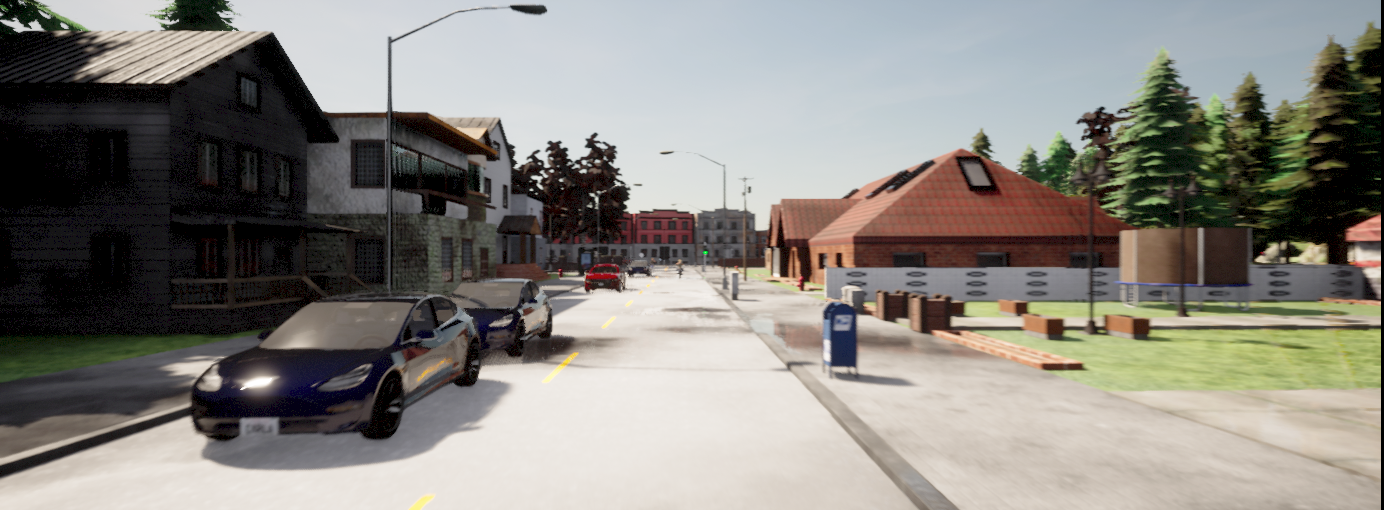}
        \caption{Origin}
        \label{fig:image1}
    \end{subfigure}
    \hfill
    \begin{subfigure}[b]{0.45\textwidth}
        \includegraphics[width=\linewidth,trim={0 0 100 50},clip]{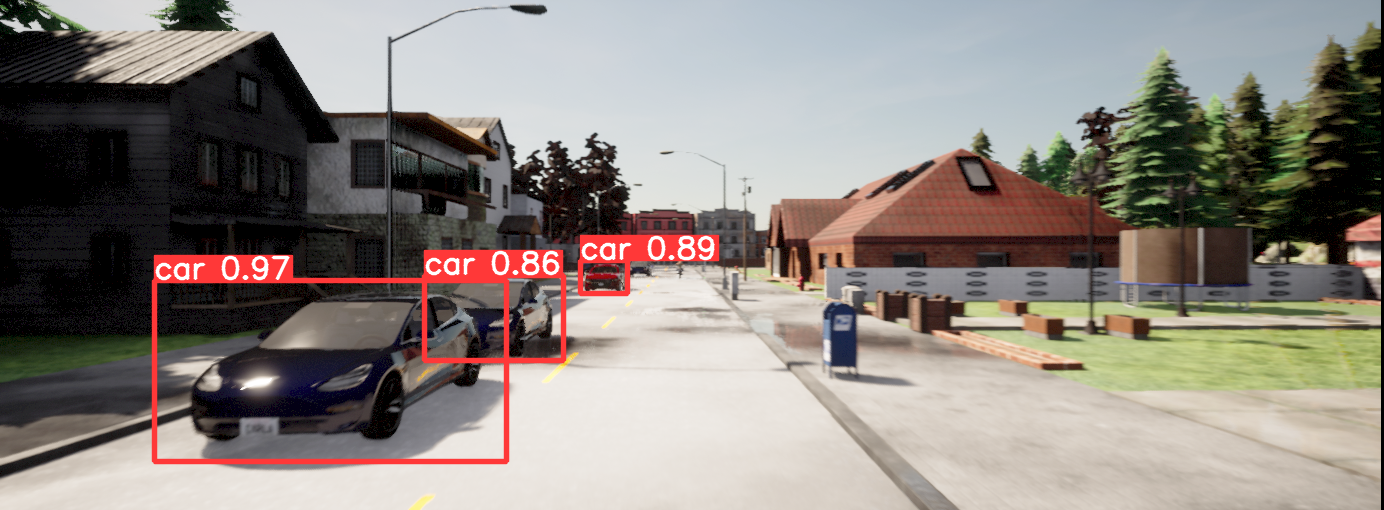}
        \caption{Active collecting}
        \label{fig:image2}
    \end{subfigure}
    \hfill
    \begin{subfigure}[b]{0.45\textwidth}
        \includegraphics[width=\linewidth,trim={0 0 100 50},clip]{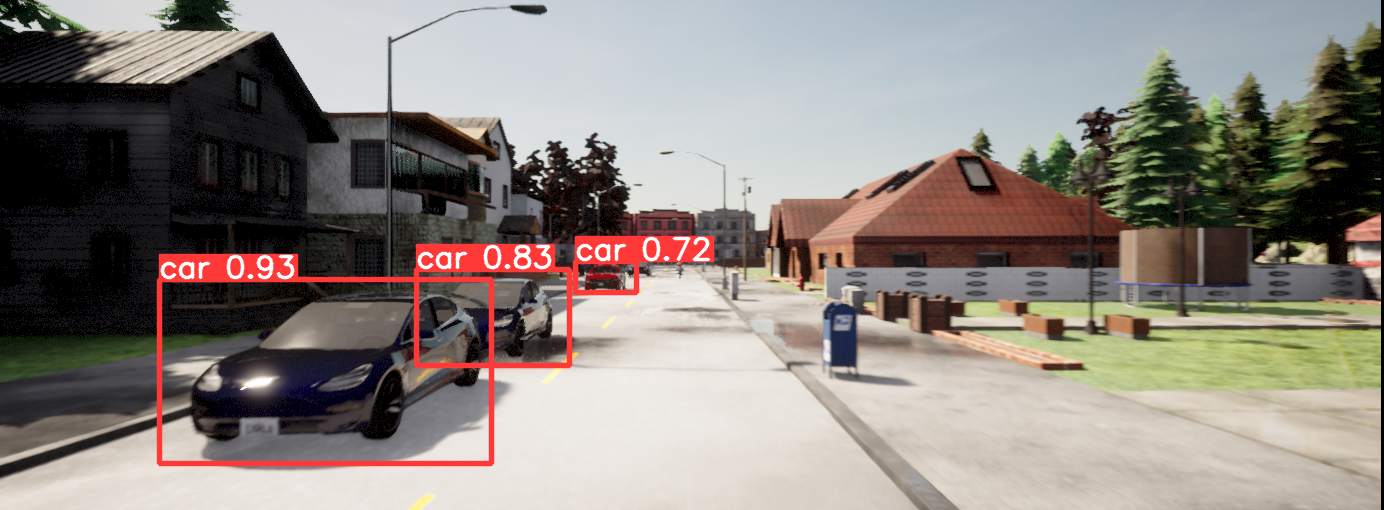}
        \caption{Passive collecting}
        \label{fig:image3}
    \end{subfigure}
    \caption{Comparison of models. Note that the confidences are different.}
    \label{fig:comparison}
\end{figure}

When dealing with images that contain multiple instances, the performance of a model trained using the active collecting strategy is generally better. This is primarily because the model has learned from a dataset that includes a diverse range of images with multiple instances within a single image.

Through this exposure, the model becomes more adept at detecting, localizing, and understanding multiple instances within a complex image. It learns to recognize patterns, relationships, and spatial configurations between different objects, enabling it to accurately identify and differentiate between the individual instances present.

This enhanced understanding of multiple instances leads to improved performance when the model is applied to real-world scenarios or tasks involving images with multiple objects. The model's ability to generalize from the diverse training examples it encountered during the active collecting process helps it better handle instances that may vary in appearance, pose, scale, or occlusion.

Furthermore, the active collecting strategy allows the model to focus on areas that are challenging or underrepresented in the dataset. This targeted selection helps address potential biases or limitations present in the initial training dataset, ensuring that the model receives sufficient exposure to instances with different characteristics, thereby reducing the risk of over-fitting.

In summary, training a model with an active collecting strategy proves advantageous when dealing with images containing multiple instances. The model's exposure to diverse examples and its improved ability to understand and differentiate between objects within a single image contribute to its superior performance in tasks related to multi-instance image analysis.

In summary, the experimental results are generally consistent with our expectations. The outcomes align with the anticipated trends and patterns we hypothesized. These findings support our initial assumptions and provide further evidence for the effectiveness and viability of the proposed methods. The observed outcomes validate the hypotheses formulated and encourage future investigations. Additional studies could explore various factors and variables to enhance the robustness and generalizability of the findings.


\section{Conclusions}
\label{sec:conclusion}

In this paper, we proposed an active data acquisition strategy for improving the performance of the algorithm of autonomous driving, which is YOLO v5. We evaluated the effectiveness of this strategy through two experiments, which demonstrated that the data collected using this strategy has a higher unit data value compared to the data collected using a passive strategy. Specifically, we found that the models trained on data collected using the active strategy achieved higher accuracy and required less time to train compared to the models trained on data collected using a passive strategy.
These results suggest that the active data acquisition strategy proposed in this work can be used to improve the performance of autonomous driving algorithms in various aspects. By collecting informative data points and reducing the data acquisition time, the proposed strategy can help to reduce the cost and time required to train autonomous driving algorithm models, while maintaining or improving the accuracy of the algorithms.
In summary, the experiments have demonstrated that the hypothesis proposed in this paper is correct. 
For the vehicle detection problem, improving the Ground Truth generation tool will enable the proposed method to be used to verify and evaluate various active parameters and corresponding model performance, thereby validating the quality of the dataset. 
Active autonomous driving data collection, from the perspective of the dataset, may become another path for further improving autonomous driving algorithms.
In the future, we will extend this work with other AI technologies, such as knowledge graph \cite{liu2022towards,lan2022semantic} and human-centered AI \cite{lan2022vision} to reduce the reality gap \cite{lan2019evolutionary,lan2019simulated}.

\section*{Acknowledgements}








\bibliographystyle{IEEEtran}
\bibliography{bibliography}

\end{document}